\documentclass[11pt]{article}

\usepackage[margin=0.88in,headheight=22pt]{geometry}
\usepackage[T1]{fontenc}
\usepackage[utf8]{inputenc}
\usepackage{lmodern}
\usepackage{microtype}
\usepackage{graphicx}
\usepackage[table,dvipsnames]{xcolor}
\IfFileExists{xurl.sty}{\usepackage{xurl}}{\usepackage{url}}
\usepackage{array}
\usepackage{booktabs}
\usepackage{tabularx}
\usepackage{ragged2e}
\usepackage{amsmath}
\usepackage{amssymb}
\usepackage{caption}
\usepackage{enumitem}
\usepackage{etoolbox}
\usepackage{float}
\usepackage{fancyhdr}
\usepackage{titlesec}
\usepackage{needspace}
\usepackage{listings}
\usepackage[most]{tcolorbox}

\definecolor{ReportInk}{HTML}{1F2933}
\definecolor{ReportMuted}{HTML}{5B6572}
\definecolor{ReportBlue}{HTML}{2563A6}
\definecolor{ReportTeal}{HTML}{138A8A}
\definecolor{ReportOrange}{HTML}{B86B17}
\definecolor{ReportPanel}{HTML}{F6F8FA}
\definecolor{ReportSoft}{HTML}{EEF7F6}
\definecolor{ReportLine}{HTML}{D7DEE8}
\definecolor{ReportCodeBg}{HTML}{F7F4EF}
\definecolor{ReportCodeFrame}{HTML}{E8D9C2}

\usepackage[
  colorlinks=true,
  linkcolor=ReportBlue,
  citecolor=ReportTeal,
  urlcolor=ReportBlue
]{hyperref}

\hypersetup{
  pdftitle={HELIX: Model--Harness Co-evolution for Recursive Self-Improvement},
  pdfauthor={Tianyu Fan, Chao Huang},
  pdfsubject={Model--harness co-evolution, auditable harness evolution, and verified sibling data},
  pdfkeywords={agent harness, model--harness co-evolution, recursive self-improvement, sibling trajectories, training data}
}

\newcommand{\system}{\textsf{HELIX}}
\newcommand{\code}[1]{\texttt{#1}}

\renewcommand{\headrulewidth}{0.4pt}
\renewcommand{\headrule}{%
  \hbox to\headwidth{\color{ReportLine}\leaders\hrule height \headrulewidth\hfill}}

\titleformat{\section}[block]
  {\Large\sffamily\bfseries\color{ReportInk}}
  {\color{ReportTeal}\thesection}{0.75em}{}
  [\vspace{0.28em}{\color{ReportTeal!35}\titlerule[0.7pt]}]
\titleformat{\subsection}
  {\large\sffamily\bfseries\color{ReportInk}}
  {\color{ReportOrange}\thesubsection}{0.7em}{}
\titleformat{\subsubsection}
  {\normalsize\sffamily\bfseries\color{ReportInk}}
  {\color{ReportTeal}\thesubsubsection}{0.65em}{}
\titlespacing*{\section}{0pt}{1.75em plus 0.25em minus 0.15em}{1.0em}
\titlespacing*{\subsection}{0pt}{1.15em plus 0.2em minus 0.1em}{0.55em}
\titlespacing*{\subsubsection}{0pt}{0.95em plus 0.15em minus 0.1em}{0.42em}
\pretocmd{\section}{\Needspace{8\baselineskip}}{}{}
\pretocmd{\subsection}{\Needspace{5\baselineskip}}{}{}
\pretocmd{\subsubsection}{\Needspace{4\baselineskip}}{}{}

\setlist[itemize]{
  leftmargin=*,
  itemsep=0.18em,
  topsep=0.2em,
  label=\textcolor{ReportTeal}{\small$\blacktriangleright$}
}
\setlist[enumerate]{
  leftmargin=*,
  itemsep=0.18em,
  topsep=0.2em,
  label=\textcolor{ReportTeal}{\arabic*.}
}

\newcolumntype{Y}{>{\RaggedRight\arraybackslash}X}
\newcommand{\tablehead}{\rowcolor{ReportTeal!10}}

\lstdefinestyle{reportcode}{
  basicstyle=\ttfamily\small\color{ReportInk},
  breaklines=true,
  columns=fullflexible,
  keepspaces=true,
  showstringspaces=false,
  keywordstyle=\color{ReportBlue}\bfseries,
  commentstyle=\color{ReportMuted}\itshape,
  stringstyle=\color{ReportOrange}
}

\newtcblisting{codeblock}[1][]{
  enhanced,
  breakable,
  listing only,
  listing options={style=reportcode},
  title={#1},
  colback=ReportCodeBg,
  colframe=ReportCodeFrame,
  coltitle=ReportInk,
  fonttitle=\sffamily\bfseries\small,
  boxrule=0.45pt,
  arc=1.5mm,
  left=2.5mm,
  right=2.5mm,
  top=1.5mm,
  bottom=1.5mm,
  borderline west={2pt}{0pt}{ReportOrange}
}

\newtcolorbox{findingbox}[1]{
  enhanced,
  breakable,
  title={#1},
  colback=ReportSoft,
  colframe=ReportTeal!45,
  coltitle=ReportInk,
  fonttitle=\sffamily\bfseries,
  boxrule=0.55pt,
  arc=2mm,
  left=4mm,
  right=4mm,
  top=2mm,
  bottom=2mm,
  borderline west={2.5pt}{0pt}{ReportTeal}
}

\newcommand{\framedgraphic}[2][0.96\linewidth]{%
  \begin{tcolorbox}[
    enhanced,
    colback=white,
    colframe=ReportLine,
    boxrule=0.45pt,
    arc=1.5mm,
    left=1.2mm,
    right=1.2mm,
    top=1.2mm,
    bottom=1.2mm,
    drop fuzzy shadow=ReportLine!55
  ]
  \centering\includegraphics[width=#1]{#2}
  \end{tcolorbox}%
}

\newcommand{\MetricPortsPerProduct}{96}
\newcommand{\MetricLcbMaxRelativeGain}{58.0\%}
\newcommand{\MetricSweCleanResolved}{205}
\newcommand{\MetricTrainingRecords}{438}

\title{\system: Model--Harness Co-evolution for \\ Recursive Self-Improvement}

\author{
Tianyu Fan, Chao Huang\\
In$^{3}$ Lab, The University of Hong Kong
}

\date{}

\makeatletter
\renewcommand{\maketitle}{%
  \thispagestyle{empty}
  \begin{tcolorbox}[
    enhanced,
    colback=ReportPanel,
    colframe=ReportLine,
    boxrule=0.45pt,
    arc=2mm,
    left=6mm,
    right=6mm,
    top=6mm,
    bottom=6mm,
    borderline west={4pt}{0pt}{ReportTeal},
    drop fuzzy shadow=ReportLine!65
  ]
    {\sffamily\bfseries\fontsize{19}{23}\selectfont\color{ReportInk}\@title\par}
    \vspace{0.75em}
    {\small\color{ReportMuted}\@author\par}
  \end{tcolorbox}
}
\makeatother

\renewenvironment{abstract}
  {\begin{tcolorbox}[
    enhanced,
    breakable,
    title={Abstract},
    colback=ReportSoft,
    colframe=ReportTeal!45,
    coltitle=ReportInk,
    fonttitle=\sffamily\bfseries,
    boxrule=0.55pt,
    arc=2mm,
    left=4mm,
    right=4mm,
    top=2mm,
    bottom=2mm
  ]\small}
  {\end{tcolorbox}}

\begin{document}

\maketitle

\vspace{0.15em}
\noindent\makebox[\linewidth][c]{%
  \includegraphics[width=0.84\linewidth]{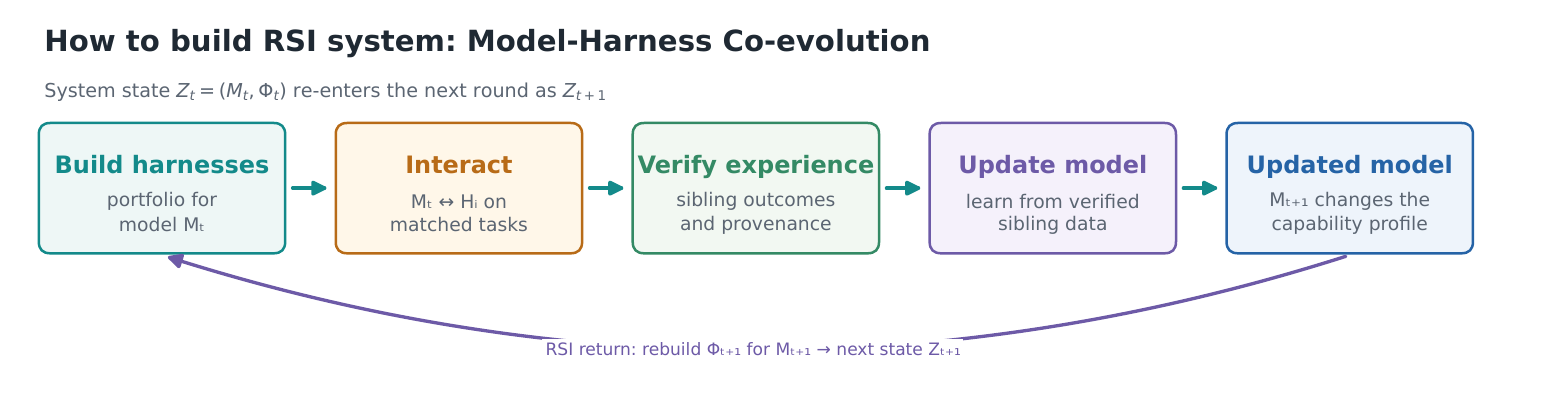}%
}

\begin{abstract}
Scaling agent capability has largely been pursued through one lens: make the model better. Yet a model acts through a runtime harness that mediates context, tools, control flow, permissions, and stopping decisions. The harness is not a neutral wrapper---it is an active co-determinant of agent behavior. For interactive agents, its influence is recursive: it governs how the model executes today and shapes the trajectories from which the model learns tomorrow. Treating model and harness as independent artifacts severs this feedback loop. We argue that closing it requires \emph{model--harness co-evolution} as a first-class principle. Co-evolution rejects a false asymmetry in which one component evolves while the other stays frozen. Harness design governs fixed-model execution and structures the data through which the model improves. A stronger model, in turn, may favor a different harness that matches its new capability profile. This bidirectional dependency gives harness building a dual purpose: producing stronger immediate execution, and generating \emph{verified sibling trajectories}---paired outcomes capturing successes, regressions, near misses, and alternative solutions---as structured training signal for the next model update.\\

Recent approaches have begun to connect harness adaptation with model learning, but making this handoff source-traceable and auditable across heterogeneous runtime designs remains difficult. We organize our approach around \emph{build--update--rebuild}: \emph{build} source-traceable harnesses for a fixed model; \emph{update} the model from verified sibling trajectories; \emph{rebuild} the harnesses for the updated model, whose shifted capability profile may favor different runtime designs. Within this paradigm, we present \system{}, a source-traceable substrate for dual-purpose harness building. It decomposes OpenCode, Pi Mono, Nanobot, and Hermes Agent into typed ports, atoms, recipes, product shells, and runtime policies---making interventions explicit rather than implicit in ad-hoc prompting. Pre-execution checks ensure auditability. An evidence plane retains traces, test results, policy decisions, and provenance, preserving outcome richness rather than collapsing to a single success bit. \system{} exposes \MetricPortsPerProduct{} ports per product contract and enumerates $4^5=1{,}024$ coupled recipes, or $4^6=4{,}096$ when acceptance is independently selected. We validate both sides of the co-evolution loop. On the execution side, a 65-candidate evolution round finds a fixed harness that improves task coverage by 4.0\% over Pi, while the complete post-hoc portfolio exposes up to \MetricLcbMaxRelativeGain{} more coverage. Selected members are then evaluated with repeated runs and the official SWE-bench evaluator. On the learning side, a 200-slot sibling slice from this deeper validation yields \MetricTrainingRecords{} verified SFT, critic, filter, and preference records. The same process that improves execution today generates the data needed for tomorrow's model update. Together, these results position source-traceable harness building as an auditable interface between execution-time harness evolution and subsequent model updating. Code is released at \url{https://github.com/HKUDS/HELIX}.
\end{abstract}

\clearpage
\tableofcontents
\clearpage

\section{Introduction}
\label{sec:introduction}

Scaling agent capability is usually framed as a model-improvement problem. Yet
a deployed agent is an executed model--runtime pair. Between a user request and
a final patch lies a substantial runtime: system instructions, context
construction, session state, tool schemas, permission checks, provider
adaptation, a turn loop, retry and compaction policies, stopping rules, and
verifiers. We call this runtime the \emph{agent harness}. It is not a neutral
wrapper. By changing what the model observes, which actions it can take, how it
recovers from failure, and when it stops, the harness co-determines the
behavior that the agent can reach.

Model and harness interact at two time scales. Within a task, the harness
constructs the model's observations and action interface; the model proposes
messages and actions; and the harness checks and executes them, returns
environment feedback, and decides whether the interaction should continue.
Across improvement rounds, verified trajectories from these interactions shape
the next model, and the changed model becomes the condition under which the
harness portfolio is rebuilt. The recursively improving unit is therefore the
model--harness system rather than the model alone: an updated pair re-enters the
same process under externally specified tasks, objectives, and verifiers.

We therefore treat \emph{model--harness co-evolution} as a first-class
principle for recursive self-improvement. Each harness-evolution round has two
outputs. The first is stronger execution for the
current model: evolution can identify a better fixed harness and expose further
coverage across a rapidly built portfolio. The second is data for the next
model update: matched successes, regressions, near misses, and alternative
solutions become \emph{verified sibling trajectories}. These outputs are not
separate goals. The same harness interventions that change execution today
structure the learning signal available tomorrow. The harness is therefore a
behavior operator for the current model and a curriculum operator for its
successor.

Recent work has already demonstrated trace-driven harness adaptation and joint
harness--model optimization \cite{harnessx,rhi}. \system{} targets a
complementary systems requirement: preserving intervention identity and
verified evidence when runtime behavior is recomposed across heterogeneous
harness families. We organize this handoff as three stages:

\begin{enumerate}
  \item \textbf{Build.} For a fixed model, construct and evaluate explicit,
  source-traceable harness candidates on matched tasks.
  \item \textbf{Update.} Turn the resulting verified sibling trajectories into
  learning records and use them to update the model.
  \item \textbf{Rebuild.} Evolve the harness again for the updated model, whose
  capability profile may prefer a different runtime design.
\end{enumerate}

Making build--update--rebuild repeatable requires more than ad-hoc prompt or
tool changes. Modern harnesses are large, tightly coupled systems, so an
intervention must remain explicit, executable, and auditable as components are
changed. \system{} provides this substrate. It decomposes OpenCode, Pi
Mono, Nanobot, and Hermes Agent into typed ports, atoms, recipes, product
shells, and runtime policies. Pre-execution checks validate declared
compositions, while an evidence layer retains traces, test results, policy
decisions, provenance, and outcome labels through execution.

Our evaluation follows one portfolio through increasing depth. A complete
65-candidate LCB evolution matrix identifies a stronger fixed harness and
exposes up to \MetricLcbMaxRelativeGain{} more post-hoc portfolio coverage than
Pi. Selected members then receive repeated-run and official SWE-bench
validation, whose sibling artifacts yield \MetricTrainingRecords{} SFT, critic,
filter, and preference records. Together, these results connect harness
evolution for the current model to the data handoff that drives the next
co-evolution round.

\subsection{Contributions}

This paper makes three contributions:

\begin{enumerate}
  \item \textbf{An auditable formulation of system-level recursive
  improvement.} We make model--harness co-evolution explicit as
  build--update--rebuild, with a task-level interaction loop and a cross-round
  return from verified experience to model updating and harness rebuilding.
  \item \textbf{A bounded harness-modification substrate.} \system{}
  represents source-derived runtime behavior as typed ports, atoms, recipes,
  product shells, and policies, then preserves each declared intervention
  through pre-execution checks, assembly, and evidence capture.
  \item \textbf{A verified experience handoff.} We show how rapid harness
  evolution exposes stronger and complementary task outcomes, and how the same
  matched rollouts materialize as SFT, critic, filter, and preference records
  for model updating.
\end{enumerate}

\section[Model--Harness Co-evolution]{Model--Harness Co-evolution through Build--Update--Rebuild}
\label{sec:coevolution}

The feedback loop introduced above becomes a repeatable process only when the
harness is treated as both an execution policy and a data-generation policy.
We treat the model--harness pair, rather than the model alone, as the state
being improved. A round becomes recursive when its verified outcomes determine
both the data used to update the model and the evidence used to rebuild the
next harness portfolio. Because tasks, objectives, and verifiers remain
externally specified, this is a bounded, verifier-grounded form of system-level
recursive improvement.

\subsection{Two coupled feedback timescales}

Let the system state at round $t$ be
$Z_t=(M_{\theta_t},\Phi_t)$, where $M_{\theta_t}$ is the current model and
$\Phi_t=\{H_{\phi_i}\}_i$ is the harness portfolio built for it. Within a
task, model and harness form a fast interaction loop: $H_{\phi_i}$ constructs
context and exposes actions, $M_{\theta_t}$ returns messages and tool calls,
and $H_{\phi_i}$ checks and executes those actions before returning the next
observation or stopping. Across rounds, a slower improvement loop converts the
resulting evidence into a model update and a rebuilt portfolio.

Let $x\sim\mathcal{D}$ be a task, $M_{\theta}$ a model with parameters
$\theta$, and $H_{\phi}$ a harness with configuration and implementation
parameters $\phi$. A run produces a trajectory
\begin{equation}
  \tau_{t,i,x,k}
  \sim
  P\!\left(
    \tau \mid x, M_{\theta_t}, H_{\phi_i}, k
  \right),
  \label{eq:trajectory}
\end{equation}
where $t$ indexes a co-evolution round, $i$ a harness candidate, and $k$ an
attempt. The trajectory includes model messages, tool calls and results,
session events, workspace effects, and stopping decisions. A verifier observes
the task, trajectory, workspace delta, test evidence, and policy evidence:
\begin{equation}
  y_{t,i,x,k}
  =
  V\!\left(
    x,\tau_{t,i,x,k},\Delta W_{t,i,x,k},
    \mathcal{T}_{t,i,x,k},\mathcal{P}_{t,i,x,k}
  \right).
  \label{eq:verifier}
\end{equation}
The label $y$ need not be binary. It may distinguish resolved, target-miss,
regression, no-action, policy violation, patch-noise, or evaluator gap.
Because $H_{\phi}$ changes both the reachable trajectory and the evidence
retained afterward, the fast interaction loop determines both current behavior
and the experience available to the slower improvement loop. A
harness-evolution round must therefore be evaluated for execution and
learning-data value together.

\subsection{Two outputs of a harness-evolution round}

For a fixed model, one deployment objective selects a harness that optimizes
verified reward while accounting for runtime cost $C$ and policy risk $P$:
\begin{equation}
  J_{\mathrm{deploy}}(M_{\theta_t},H_\phi)
  =
  \mathbb{E}_{x,k}
  \left[
    R(y,\tau)
    -\lambda C(\tau)
    -\mu P(\tau)
  \right].
  \label{eq:harness-objective}
\end{equation}
The weights $\lambda$ and $\mu$ make explicit that ``solves more tasks'' is
not the only deployment objective. A harness that succeeds through excessive
tool calls, unsafe permissions, or flaky retries can be inferior to a slightly
less accurate but predictable one.

Fixed-harness utility is only the first output. Holding model and task fixed,
the evaluated candidates also create a sibling set
\begin{equation}
  \mathcal{S}_{t,x}
  =
  \left\{
    \left(
      H_{\phi_i},\tau_{t,i,x,k},y_{t,i,x,k}
    \right)
  \right\}_{i,k}.
  \label{eq:siblings}
\end{equation}
This matched grouping is more informative than an undifferentiated replay
buffer. A resolved patch paired with a no-action sibling isolates a failure to
turn analysis into execution. Two resolved siblings with different patch
hygiene create a minimality preference. A target-passing patch with a
regression creates a critic negative that cannot be identified from the target
test alone.

A harness-evolution round therefore evaluates a \emph{portfolio} rather than
only its eventual winner. Let $Q_{\mathrm{data}}$ reward verified task novelty,
contrasting outcomes, semantic proximity, and artifact quality among the
trajectories produced by a candidate set $\mathcal{C}$, and let
$\mathcal{S}_t(\mathcal{C})$ denote its task-indexed sibling groups. A
dual-purpose objective can be written schematically as
\begin{equation}
  \mathcal{C}_t^\star
  =
  \arg\max_{\mathcal{C}\subseteq\mathcal{H}_t}
  \left[
    \alpha\max_{H\in\mathcal{C}}J_{\mathrm{deploy}}(M_{\theta_t},H)
    +\beta Q_{\mathrm{data}}(\mathcal{S}_t(\mathcal{C}))
    -\gamma C_{\mathrm{build}}(\mathcal{C})
  \right].
  \label{eq:dual-objective}
\end{equation}
The first term asks which fixed harness is useful now. The second asks what the
full candidate portfolio reveals for learning. The objectives need not select the same
candidate: a losing harness can supply a clean critic negative, while two
successful harnesses can supply a minimality preference. This paper does not
collapse the terms into one tuned scalar; it measures their observable proxies
separately. 

\subsection{From verified experience to the next system state}

Let $E_t=\{\mathcal{S}_{t,x}\}_{x\in\mathcal{D}}$ denote the verified,
task-indexed evidence from a round. A dataset builder $G$ maps this evidence to
training records, and an update operator $U$ changes the model:
\begin{equation}
  D_t = G(E_t),
  \qquad
  \theta_{t+1}
  =
  U(\theta_t,D_t).
  \label{eq:model-update}
\end{equation}
The update operator may implement supervised fine-tuning, preference
optimization, verifier or critic training, reinforcement learning, or
router/reranker learning. \system{} supplies the verified sibling inputs through
rollout execution, verification, and evidence capture. A task-specific exporter
instantiates $G$ over curated sibling groups.

The same evidence also returns to the harness side. It can guide which
component combinations are retained, pruned, or proposed when the portfolio is
rebuilt for the updated model:
\begin{equation}
  \Phi_{t+1}
  =
  B(\Phi_t,E_t;M_{\theta_{t+1}}),
  \qquad
  Z_{t+1}=(M_{\theta_{t+1}},\Phi_{t+1})=\mathcal{R}(Z_t).
  \label{eq:system-update}
\end{equation}
Here $U$ and $B$ need not use the same records: $D_t$ carries supervision for
the model, while component-level successes, failures, and incompatibilities in
$E_t$ provide evidence for the next build. The return to $Z_{t+1}$ is what
distinguishes improvement recursion from a one-off runtime refinement.

The rebuild step is necessary because harness utility is conditional on the
model: the relevant objective is $J(M_\theta,H_\phi)$. Model updates can change
error recovery, schema adherence, context sensitivity, stopping behavior, and
the value of specific tools. The equality
\begin{equation}
  \arg\max_{\phi}J(M_{\theta_{t+1}},H_\phi)
  =
  \arg\max_{\phi}J(M_{\theta_t},H_\phi)
\end{equation}
need not hold.
Consequently, permanently attaching a harness built for $M_{\theta_t}$ to
$M_{\theta_{t+1}}$ can leave performance on the table. Conversely, training a
model on trajectories from only one harness can over-specialize it to that
harness's tool vocabulary and termination behavior. Rebuilding after each
model update makes both dependencies observable.

\begin{figure}[H]
  \centering
  \framedgraphic{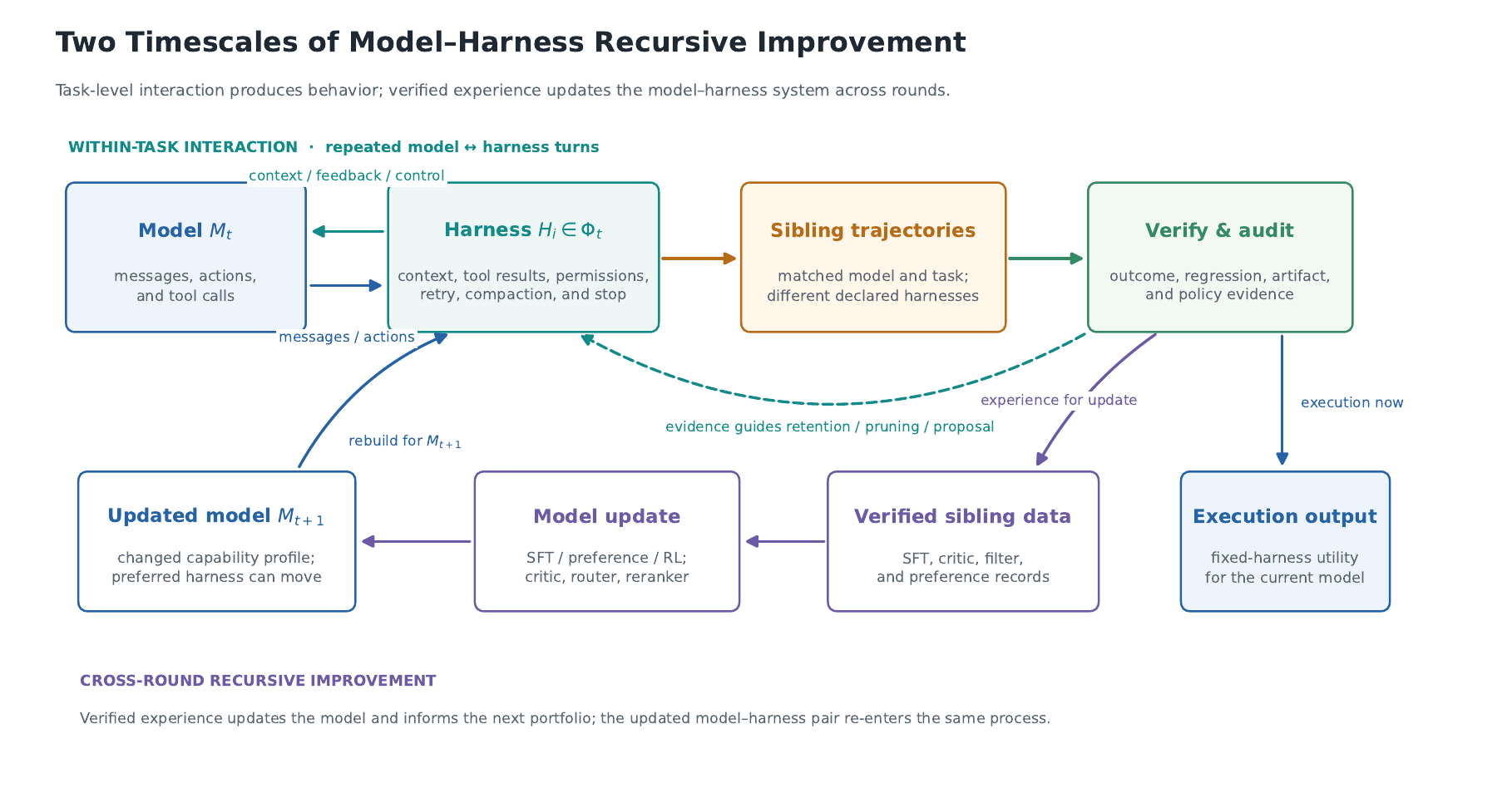}
  \caption{\textbf{Two timescales of model--harness recursive improvement.}
  Within a task, $H_i$ supplies context, tool results, permissions, and control
  decisions to $M_t$, while $M_t$ returns messages and actions. Across rounds,
  verified sibling trajectories supply the update signal for $M_{t+1}$, whose
  changed capability profile conditions the next harness rebuild.}
  \label{fig:coevolution}
\end{figure}

Figure~\ref{fig:coevolution} separates the two timescales. Within a task, model
and harness alternate between mediated observations and actions. Across rounds,
one evidence branch reports execution for the current system, while the other
organizes sibling contrasts for the update and subsequent rebuild.

\subsection{Fixed-harness performance versus portfolio yield}
\label{sec:metrics}

Fixed-harness performance and portfolio breadth require different reported
quantities. For a verifier $v\in\{0,1\}$, the observed fixed-harness task
coverage with $K$ attempts is
\begin{equation}
  \operatorname{Cov}_{K}(H)
  =
  \frac{1}{|\mathcal{D}|}
  \sum_{x\in\mathcal{D}}
  \mathbb{1}
  \left[
    \max_{1\leq k\leq K}
    v(\tau_{H,x,k})=1
  \right].
  \label{eq:fixed-coverage}
\end{equation}
For a candidate set $\mathcal{C}$ produced in a build round, post-hoc oracle coverage is
\begin{equation}
  \operatorname{OracleCov}_{K}(\mathcal{C})
  =
  \frac{1}{|\mathcal{D}|}
  \sum_{x\in\mathcal{D}}
  \mathbb{1}
  \left[
    \max_{\substack{H\in\mathcal{C}\\1\leq k\leq K}}
    v(\tau_{H,x,k})=1
  \right].
  \label{eq:oracle-coverage}
\end{equation}
Equation~\ref{eq:oracle-coverage} measures potential positive-outcome yield or
the routing ceiling of an observed candidate portfolio. It is deployable only if a router can choose
the successful candidate before seeing the answer. We therefore report fixed
coverage, successful attempt slots, and oracle coverage as separate quantities.

With current execution separated from portfolio yield, we next describe how
\system{} builds explicit harness interventions and preserves the evidence
needed for both outputs.

\section[HELIX System]{\system{}: Explicit and Auditable Harness Evolution}
\label{sec:system}

\system{} operationalizes the harness side of build--update--rebuild. It turns
a monolithic runtime into explicit, source-traceable interventions whose
identities persist through declaration, compilation, execution, and evidence
capture. This section describes the representation, pre-execution checks,
runtime, and evidence layer that make both build and rebuild auditable.

\subsection{The harness as an explicit intervention variable}

\system{} represents a harness as a structured composition rather than a
single product label:
\begin{equation}
\begin{split}
H=\big(&H_{\mathrm{shell}},
H_{\mathrm{session/hooks}},
H_{\mathrm{config}},
H_{\mathrm{prompt}},
H_{\mathrm{tools}},\\
&H_{\mathrm{turn}},
H_{\mathrm{accept}},
H_{\mathrm{policy}}\big).
\end{split}
\label{eq:harness-tuple}
\end{equation}
These eight dimensions cover both the model-facing interface and the
environment-facing control plane. The shell exposes SDK, CLI, TUI, or web
entrypoints; CLI-Anything provides a concrete example of an agent-facing CLI
harness for existing software \cite{cli-anything}. Session and hooks control
lifecycle and extension events. Config normalizes provider and product
behavior. Prompt and tools define the model's observations and action
vocabulary. The turn loop governs provider/tool iteration, continuation,
retry, and compaction. Acceptance determines what evidence is sufficient to
stop. Runtime policy constrains permissions and execution.

\subsection{Source-traceable ports, atoms, and recipes}

\system{} uses six entities to preserve intervention identity. A \emph{port}
defines a stable capability slot. \emph{Common atoms} provide product-neutral
behavior, while \emph{personality atoms} preserve source-specific semantics
behind the same ports. A \emph{pack} expands a convenient group of atoms. A
\emph{recipe} binds atoms, strategies, policies, and shells into one declared
candidate, and a \emph{product shell} exposes that candidate through an SDK,
CLI, TUI, or web surface. The compiler and lockfile retain the full expansion
used by each rollout.

The contracts derive behavior from OpenCode, Pi Mono, Nanobot, and Hermes
Agent \cite{opencode-repo,pi-repo,nanobot-repo,hermes-repo}. Each full product
contract exposes \MetricPortsPerProduct{} ports, bindings, and swap points.
The same interfaces also support a neutral \emph{Minimal} contract used for
compact runtime and conformance checks.

\subsection{Pre-execution checks and recipe enumeration}
The recipe compiler expands a declared composition, resolves dependencies and
port bindings, and emits a deterministic lockfile for runtime assembly.
Conformance, boundary, and source-purity checks flag invalid or ambiguous
compositions before live evaluation. Together, these steps make candidate
interventions executable and auditable without treating arbitrary modules as
semantically interchangeable. Coupling session with hooks and acceptance with
the turn loop leaves five independent source choices, defining
$4^5=1{,}024$ candidates; selecting acceptance independently yields
$4^6=4{,}096$. We enumerate this finite intervention space and use deterministic
smoke screening to select candidates for live evaluation. These operations
support repeatable build and rebuild while keeping every evaluated rollout
linked to its intended intervention. The live experiment evaluates the
structured 65-candidate slice defined in Section~\ref{sec:lcb-method}, not all
1,024 or 4,096 possible recipes.

\begin{figure}[t]
  \centering
  \framedgraphic{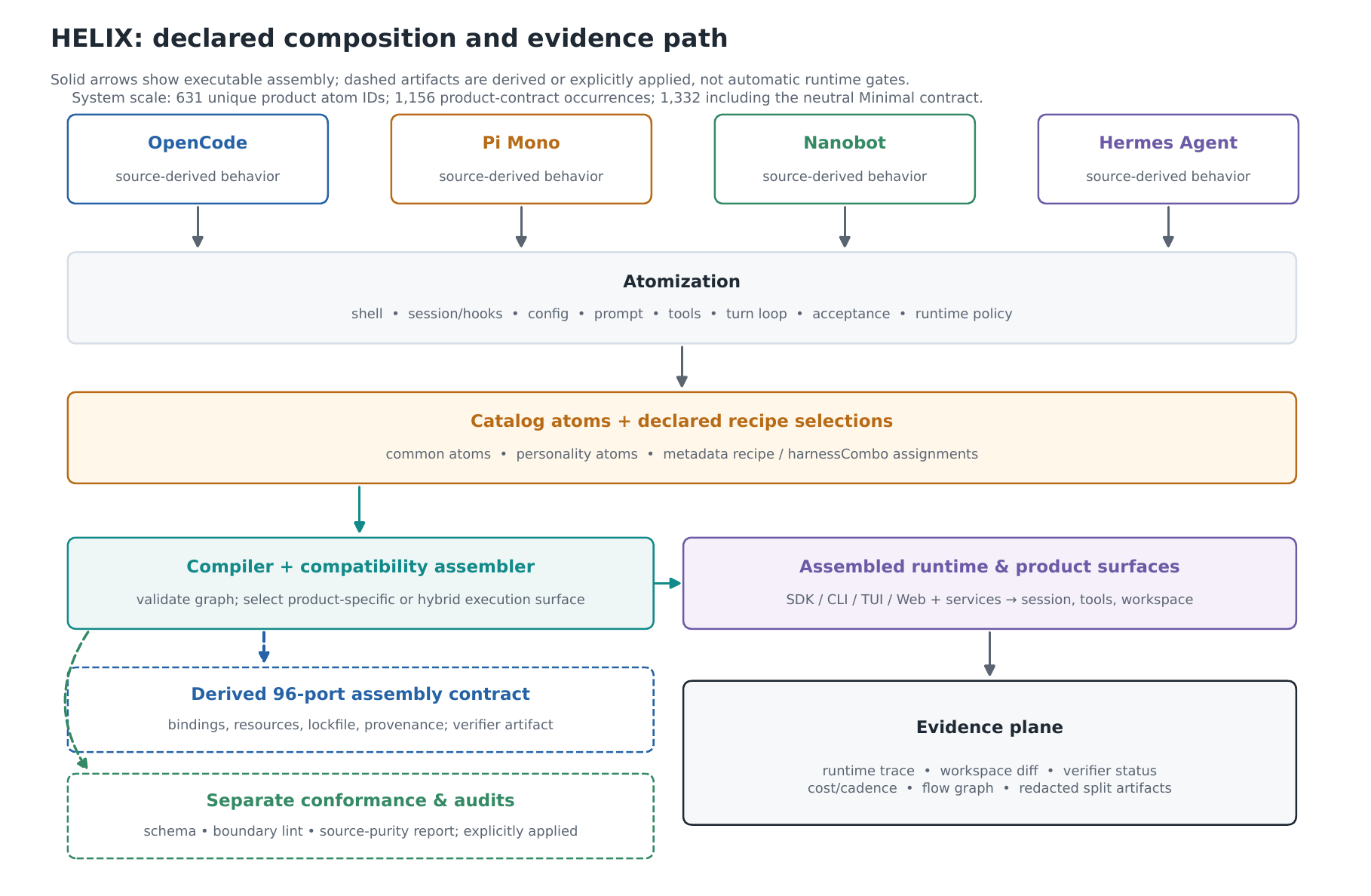}
  \caption{\textbf{\system{} architecture.} Source-derived behavior is
  represented behind stable ports and assembled into declared runtime
  candidates. Solid arrows show executable assembly; dashed paths show derived
  or explicitly applied evidence. The figure distinguishes unique atom
  identifiers from contract-expanded occurrences.}
  \label{fig:architecture}
\end{figure}

Figure~\ref{fig:architecture} summarizes the declared composition model, audit
commands, compatibility assembly path, and evidence layer.

\subsection{Model--harness interaction at runtime}

Each trajectory alternates between harness mediation and model action. The
harness constructs the current observation and action interface, the model
proposes a response, and the harness checks and executes its actions before
returning environment feedback. The assembled runtime implements this
interaction as a shared multi-step loop:

\begin{enumerate}
  \item create or restore the session, run input and before-start hooks, and
  persist the user message;
  \item normalize the request and build prompt/context under the selected
  product personality;
  \item stream a provider step, normalize assistant parts, and collect proposed
  tool calls;
  \item run tool preflight and permission checks, execute serial or parallel
  batches, apply result hooks, and truncate or retry where required;
  \item evaluate acceptance evidence and the provider finish reason;
  \item inject a continuation when required, or stop at acceptance, request
  boundary, or the configured step limit; and
  \item persist the transcript and session state, return the runtime trace, and
  let the task-parity/evaluation layer snapshot workspace effects.
\end{enumerate}

Context compaction, tool input normalization, permission strategy, continuation
cadence, and assistant-part protocols can all change a trajectory without
changing model weights. Conversely, the model's messages and actions determine
which harness transitions, tools, and stopping checks are reached next. This
bidirectional interaction is the concrete mechanism behind the coupled
distribution in Equation~\ref{eq:trajectory}: neither component produces the
observed trajectory in isolation.

\subsection{The evidence plane preserves intervention identity}

Each run is joined to its recipe and lockfile, logged model--task--attempt
identity, runtime trace, workspace effects, verifier outcome, and policy
evidence. The resulting envelope keeps the declared intervention attached to
every performance comparison and exported record.

\begin{findingbox}{HELIX makes harness evolution explicit and auditable}
Declared recipes become executable interventions whose identities persist from
compilation through runtime evidence. The same identity supports fixed-runtime
comparison and matched sibling-data construction, providing the build-to-update
handoff required by model--harness co-evolution.
\end{findingbox}

\section[Design Benefits]{Design Capabilities for Auditable Recursive Improvement}
\label{sec:data}

Section~\ref{sec:system} described how \system{} represents, assembles, and
executes harness interventions. The value of this design is not merely that it
produces more candidates. It changes the unit at which a harness can evolve,
keeps each change attached to the evidence it produces, and turns the resulting
comparisons into model-update data. 
From an RSI perspective, these mechanisms provide three capabilities needed to
repeat the cycle: bounded modification of runtime behavior, attributable
observation of its effects, and transfer of verified experience into the next
update.

\subsection{Bounded modification makes build and rebuild practical}

Editing a harness couples many concerns: changing a tool interface
can also disturb session behavior, provider adaptation, stopping rules, or the
product shell. Repeating that process for every candidate makes both initial
construction and later rebuilding expensive engineering operations. \system{}
instead makes the recipe the unit of evolution. Source-derived atoms preserve
existing behavior behind stable ports, while the compiler and shared execution
and evidence paths are reused across candidates.

The resulting design space is both broad and meaningful. A candidate selects
known behavior families at declared capability slots rather than combining
anonymous code fragments or duplicating a complete runtime. After a model
update, the same contracts can be reused while recipes are revised for the new
capability profile. Build and rebuild therefore become repeatable
component-level operations, making rapid portfolio construction feasible
without treating every candidate as a new agent implementation. For recursive
improvement, this is a bounded modification surface: the system can revise its
runtime policy through declared substitutions without relying on unconstrained
self-rewriting.

\subsection{Persistent identity enables cross-round attribution}

A score is difficult to interpret when a candidate name hides the configuration
and code that actually ran. Wrapper leakage, implicit defaults, or an invalid
binding can make the executed harness differ from the intended intervention.
\system{} carries the expanded recipe and lockfile through compatibility
assembly and joins them to the model--task--attempt identity and runtime
evidence. Conformance, boundary, and source-purity checks expose undeclared
behavior that would otherwise silently change the candidate's meaning.

This persistent identity makes an outcome useful beyond a leaderboard row. The
declared composition and logged execution conditions can be reconstructed, a gain or
regression can be associated with an inspectable component set, and
incompatibilities can guide the next round of candidate construction. The
result is not simply ``version A beats version B,'' but evidence about a
declared harness design that can be deployed, diagnosed, or rebuilt.
Across rounds, the same identity also separates two otherwise confounded
questions: how an updated model behaves under a held harness, and how rebuilt
harnesses change behavior for that updated model. Recipe, model, task, and
attempt identity therefore provide the lineage needed to attribute recursive
changes to the model, the harness, or their interaction.

\subsection{Evidence-linked siblings carry experience into model updating}

Harness modification produces trajectories, but trajectories alone are not yet
structured learning data. Runs from unrelated configurations are difficult to
compare when their intervention identity, workspace effects, or evaluation
evidence has been lost. \system{} retains these fields in one evidence envelope
and groups runs for the same model and task across declared harness
interventions. Matched sibling groups are therefore produced by construction
rather than recovered afterward from heterogeneous logs.

This structure increases the learning value of each evolution round. A sibling
group can distinguish a clean resolution from a regression, a semantic near
miss, a no-action run, or a noisier alternative solution while preserving which
harness produced each outcome. Table~\ref{tab:data-mapping} summarizes how the
resulting contrasts support different model-update objectives.
The portfolio thus acts as a curriculum operator for its current model: it
elicits multiple behaviors on matched tasks, while external execution evidence
grounds which of those self-generated experiences should be learned from,
contrasted, or filtered.

\begin{table}[H]
\centering
\small
\begin{tabularx}{\linewidth}{
  >{\RaggedRight\arraybackslash}p{0.24\linewidth}
  >{\RaggedRight\arraybackslash}p{0.28\linewidth}
  Y}
\toprule
\tablehead Verified pattern & Export & Data Description \\
\midrule
Resolved, path-clean, minimal patch &
Clean SFT example &
A concise successful trajectory/trace. \\
Resolved versus regression, target miss, or no action &
Critic example; outcome preference &
why local progress is not sufficient for task resolution. \\
Clean versus noisy resolved siblings &
Filter example/preference pair for a certain scenario&
Separate solution correctness from artifact quality. \\
Different clean resolutions &
Alternative SFT examples; minimality/cost preference &
Preserve solution diversity while preferring a smaller or cheaper trajectory. \\
\bottomrule
\end{tabularx}
\caption{\textbf{Evidence-linked siblings support multiple learning
objectives.} The benefit is not a particular export format: intervention
identity and outcome evidence remain attached as the same rollouts are reused
for SFT, critic, filter, or preference learning.}
\label{tab:data-mapping}
\end{table}

Persistent task and sibling identity also allows the exporter to keep every
sibling group in one train/dev split, preventing paired or near-identical
trajectories from leaking across the split boundary.

Together, these capabilities connect all three stages of co-evolution. Bounded
component reuse makes build and rebuild repeatable; persistent identity makes
cross-round effects attributable; and evidence-linked siblings turn the
current system's interaction experience into provenance-preserving data for
the update stage.
We next evaluate the execution and learning outputs enabled by this design.

\Needspace{14\baselineskip}
\section[Experimental Design]{Experimental Design: Two Outputs of Fixed-Model Harness Evolution}
\label{sec:method}

\subsection{Evaluation goals and research questions}

We evaluate one 65-candidate evolution portfolio under a shared model label,
then deepen the evidence for selected members through repeated-run,
cross-benchmark, and artifact-quality analysis.

We therefore ask these three research questions:

\begin{itemize}
  \item \textbf{RQ1: Execution beyond the baseline.} Under a shared
  model label, can evolution find a source-pure fixed harness that surpasses
  the baseline, and what happens when selected members receive deeper
  evaluation?
  \item \textbf{RQ2: Execution breadth from rapid evolution.} Beyond the best
  fixed harness, what additional task coverage and sibling diversity does the
  complete evolved portfolio expose?
  \item \textbf{RQ3: Learning data for the next update.} What diversity of
  verified sibling outcomes does evolution produce, and how can
  selected-member validation turn its more deeply labeled sibling trajectories
  into SFT, critic, filter, and preference records?
\end{itemize}

RQ1 asks whether the evolved population contains a stronger fixed runtime.
RQ2 keeps the whole population and measures the behavior exposed beyond that
winner. RQ3 follows selected siblings through stronger outcome and patch-quality
labels into model-update data.

\subsection{LiveCodeBench-derived repair subset}
\label{sec:lcb-method}

The main evolution evaluation uses 100 fixtures derived from
\code{livecodebench/code\_generation\_lite}
\cite{livecodebench}: 26 easy, 24 medium, and 50 hard. Each fixture asks the
agent to modify a local \code{solution.py}. All LCB-derived results use
\emph{trace-strict status}: a successful slot requires an observed normalized
bash call that executes \path{test_solution.py} with \code{python} or
\code{python3}, together with the corresponding passing output.

The main evolution matrix contains Pi Mono and 64 Pi-centered source-pure
recipes. The recipes enumerate the $4^3$ choices for session/hooks, config, and
tools while retaining the Pi prompt, turn loop, and acceptance behavior. Every
one of the 65 candidates runs once on every task, producing 6,500 slots. This
complete matrix is the primary evidence for fixed-candidate comparison and
portfolio breadth.

A repeated-run follow-up selects Pi and two OpenCode-family members from the
same 65-candidate portfolio. These source-pure interventions were chosen for
deeper mechanism analysis, not as the top two candidates in the one-attempt
matrix. Each receives ten attempts per task, producing 3,000 slots. The
follow-up measures both tasks with at least one success and successful slots,
which capture repeated-run reliability. Detailed rows are reported in
Appendix~\ref{app:selected-results}.

\subsection{Selected-member SWE-bench Verified follow-up}
\label{sec:swe-method}

The cross-benchmark follow-up samples 55 instances from
SWE-bench~\cite{swebench}. It transfers selected members of the same portfolio:
Pi, the two leading fixed candidates in the one-attempt LCB matrix, and the two
OpenCode-family members used in the repeated-run follow-up. Each receives two
attempts per instance, producing 550 slots. The maximum is 200 agent steps,
local concurrency is one, and task resolution is labeled by the official
SWE-bench evaluator. This matrix supplies the official outcomes, patches, and
sibling traces used for data analysis in RQ3.

\subsection{Shared model and attempt protocol}

Run metadata identifies \code{MiniMax-M2.7-highspeed} with
\code{deterministic=false} for all SWE-bench and LiveCodeBench rollouts.

\section[Results]{Results: Execution Today and Learning Data Tomorrow}
\label{sec:results}

The results follow one evidence chain. We first analyze all 65 candidates, then
use selected-member follow-ups to test repeated-run behavior, transfer to an
official evaluator, and deepen the labels needed for data materialization.

\subsection{RQ1---Execution today: can evolution surpass the baseline?}

\paragraph{Complete evolution matrix.}
Across all 65 candidates, Pi solves 50/100 tasks. The best and second-best fixed
candidates solve 52/100 and 51/100, respectively. The winner uses Hermes
session/hooks with the remaining components from Pi; the runner-up combines
Hermes session/hooks and config with Nanobot tools and the Pi core. The best
fixed result is a 4.0\% relative improvement over Pi, so RQ1 has a positive but
narrow answer: evolution finds a stronger fixed harness, but only two of 64
non-baseline candidates exceed Pi.

\begin{figure}[H]
  \centering
  \framedgraphic{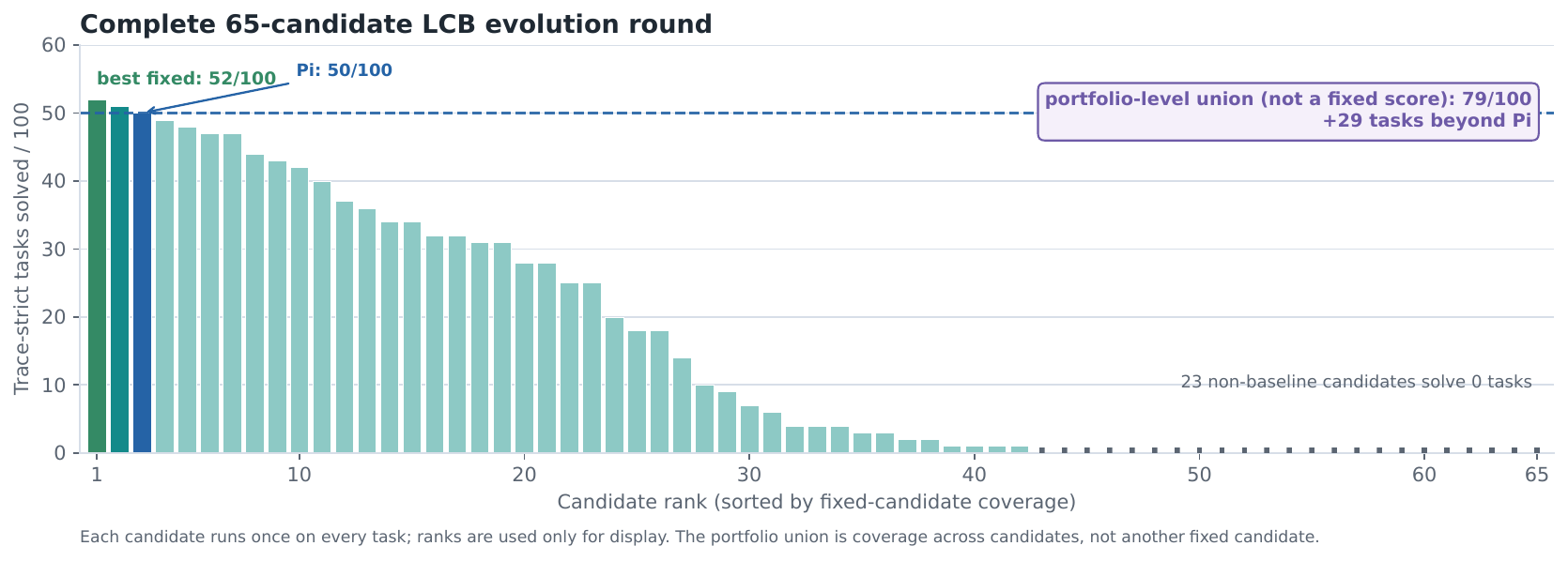}
  \caption{\textbf{The complete 65-candidate distribution is shown.} Bars show
  trace-strict fixed-candidate coverage under one attempt per task and are
  sorted only for display. Pi solves 50/100 tasks, two candidates exceed it,
  and the 23 zero-score non-baseline candidates are marked along the axis. The
  79/100 post-hoc union is portfolio coverage, not the score of another fixed
  candidate.}
  \label{fig:benchmark-results}
\end{figure}

\paragraph{Selected-member validation.}
In the repeated LCB follow-up, the selected OpenCode-family members each cover
75/100 tasks versus
72/100 for Pi. Their successful-slot totals are 549 and 546 versus 529 for Pi.
Coverage records whether a task succeeds at least once, whereas slots measure
reliability across all attempts. Although the attempt budgets differ, these
rows deepen the evidence rather than rerank the one-attempt matrix.

On SWE-bench, the selected-member follow-up changes the preferred fixed
harness. The two candidates that lead the one-attempt LCB matrix each resolve
44/55 instances, matching Pi, while the selected OpenCode-family members
resolve 46/55. The strongest selected member records 86 resolved slots versus
81 for Pi. Exact candidate rows for both follow-ups appear in
Appendix~\ref{app:selected-results}.

\begin{findingbox}{Finding 1: harness evolution improves execution today}
The complete 65-candidate matrix contains a fixed harness that surpasses Pi
(52/100 versus 50/100). Selected-member follow-ups also contain stronger fixed
harnesses under repeated LCB evaluation (75/100 versus 72/100) and official
SWE-bench evaluation (46/55 versus 44/55), but the preferred member changes
with the protocol. Harness evolution can therefore improve execution without
implying one universal winner across environments.
\end{findingbox}

Finding 1 establishes the first execution result: evolution can produce a
stronger fixed runtime for the current model. We next retain the wider evolved
portfolio and ask how much additional behavior becomes reachable beyond that
single candidate.

\subsection{RQ2---Execution breadth from rapid harness evolution}

The complete 65-candidate portfolio covers 79/100 tasks, 29 more than the Pi
row in the same one-attempt matrix, and contains 1,014 trace-strict successes
across 6,500 slots. This 58.0\% relative coverage gain is a post-hoc union over
the evolved portfolio, not a deployable fixed-harness score. Its value is the
additional behavior and sibling outcomes exposed by rapid evolution.

The same matrix also shows why evolution requires selection. Only two of 64
non-baseline candidates exceed Pi, while 23 solve no task. Blind composition is
therefore a poor deployment strategy; typed gates, cheap screens, adaptive
selection, and eventually a learned router are needed to exploit
complementarity without paying for every candidate.

The selected-member SWE follow-up provides an official-evaluator check of the
same phenomenon. Its post-hoc union covers 49/55 instances, compared with
46/55 for the best fixed member and 44/55 for Pi. This is additional evidence
of complementarity and potential positive-data yield, not an online routing
result.

\begin{findingbox}{Finding 2: rapid evolution expands execution and sibling diversity}
\textbf{Evolution requires selection.} Only two of 64 evolved LCB candidates
outperform Pi and 23 solve no task. Arbitrary component composition is
therefore unreliable; typed gates, cheap screening, and adaptive selection are
necessary parts of harness evolution.

\textbf{Evolution creates complementary coverage and data.} The best fixed
LCB candidate moves from 50/100 to 52/100, whereas the complete portfolio
covers 79/100. The selected-member SWE union similarly exceeds its best fixed
member, 49/55 versus 46/55. These post-hoc gaps are not deployable scores. They
show that rapid evolution uncovers task-specific successes and failures that
can support routing and provide diverse sibling data for model updating.
\end{findingbox}

The additional outcomes in Finding 2 are the raw material for the learning
side of co-evolution. RQ3 asks which sibling contrasts become structured
records for the next model update.

\subsection{RQ3---Learning data for the next model update}

\paragraph{From portfolio outcomes to usable labels.}

The 65-candidate matrix supplies broad trace-strict outcome diversity, but
additional coverage is only potential data yield. A rollout becomes a
plausible learning record only after outcome validity, regression behavior,
patch hygiene, and intervention identity are examined. The selected-member SWE
validation provides these deeper labels. Across its 550 evaluated slots, 543
contain nonempty patches. Official statuses are 411 resolved, 132 gaps, and
seven empty patches. Of the nonempty patches, 486
modify production code only, 57 touch tests, and 248 contain symlink/path
noise. Among the 411 resolved slots, 363 are production-only, but only
\MetricSweCleanResolved{} are both production-only and path-clean; 48 resolved
patches touch tests. Official resolution is therefore necessary but not
sufficient for a clean training positive.

\paragraph{Sibling contrasts reveal supervision.}
\label{sec:cases}

The aggregate funnel measures quantity; paired cases identify the supervision
carried by that yield. We draw three SWE-bench contrasts from the
official-evaluator follow-up and two LCB-derived contrasts from the repeated-run
follow-up. The former are the primary data cases; the latter provide supporting evidence about repair closure
and harness-evolution fragility. Table~\ref{tab:selected-cases} states the selection logic
before interpretation. Appendix~\ref{app:case-traces} provides the
verifier-grounded traces for the Pytest and Xarray contrasts.

\begin{table}[H]
\centering
\footnotesize
\begin{tabularx}{\linewidth}{
  >{\RaggedRight\arraybackslash}p{0.21\linewidth}
  >{\RaggedRight\arraybackslash}p{0.29\linewidth}
  Y}
\toprule
\tablehead Case & Matched contrast & Why it is useful \\
\midrule
SWE \code{sphinx-10673} &
Pi 0/2; other siblings 7/8 resolved; selected gap patch 6,661 bytes vs.\
resolved patch 758 bytes &
A baseline-miss/sibling-hit contrast that links candidate complementarity to a
clean official-evaluator positive. \\
SWE \code{pytest-5787} &
Matrix 2 resolved / 8 gaps; selected Pi patch passes 2/2 F2P but fails two P2P
tests, selected sibling passes both sets &
Shows why target-test success alone is an unsafe positive and supplies a
regression-aware critic/preference pair. \\
SWE \code{xarray-6461} &
Matrix 9 resolved / 1 gap; same OpenCode shell/tools and attempt index, Pi
config yields an 837-byte gap while OpenCode config yields a 778-byte resolve &
Both patches are single-file, production-only, path-clean, and regression-free;
the observed hygiene is held constant while semantic outcomes differ. \\
LCB \code{v6-025-abc391-f} &
Trace-strict: Pi 0/10 vs.\ OpenCode session/hooks/config/tools 4/10; first
candidate success at attempt 2 &
Representative candidate repairs after two failed bash calls, rewrites, then
executes a passing verifier: a supporting closure path. \\
LCB \code{v6-088-arc191-d} &
Trace-strict: Pi 7/10 vs.\ the two recompositions 2/10 and 4/10 &
Both substitutions regress on the same task, preventing a monotonic
``more active harness is better'' story. \\
\bottomrule
\end{tabularx}
\caption{\textbf{Cases drawn from selected-member validation matrices.}
Selection deliberately includes gains, losses, semantic near misses, and
regression evidence. Counts are descriptive outcomes from non-deterministic
rollouts, not single-factor causal estimates.}
\label{tab:selected-cases}
\end{table}

\paragraph{Sphinx: harness evolution yields a new verified positive.}
For \code{sphinx-doc\_\_sphinx-10673}, both Pi attempts miss the target while
seven of the other eight sibling slots resolve. The selected 758-byte
OpenCode-config patch handles Sphinx's virtual index documents directly in
\code{TocTree.parse\_content()}, passes the target test, and preserves all
regression tests. The selected 6,661-byte Pi patch changes three files and
still misses the target. The case illustrates that a broader patch is not
necessarily closer to the correct semantic basin. It is the clearest
end-to-end example of candidate complementarity becoming a clean positive and a
resolved-versus-miss preference contrast.

\paragraph{Pytest: target success with regressions is still a negative.}
For \code{pytest-dev\_\_pytest-5787}, the selected Pi patch passes both
FAIL\_TO\_PASS tests but breaks
\path{test_xdist_longrepr_to_str_issue_241} and
\path{test_deserialization_failure}. The selected OpenCode-config sibling
passes all 123 PASS\_TO\_PASS tests as well as both targets. The official
evaluator therefore labels the former \code{gaps-found} and the latter
\code{resolved}. This is a high-value critic negative: target-only evaluation
would assign the wrong label. Appendix~\ref{app:trace-pytest} traces the paired
inspection, patch, validation, and official-evaluator outcomes in detail.

\paragraph{Xarray: clean does not imply correct.}
For \code{pydata\_\_xarray-6461}, the sole gap and selected resolved sibling
are both clean by patch-source criteria. The gap falls back to the condition's
attributes when the scalar \code{x} input has no attribute dictionary; the
resolved patch falls back to an empty dictionary, preserving the requested
\code{keep\_attrs} semantics. Since the provider is non-deterministic and each
candidate has only two attempts, this case illustrates a configuration-aligned
semantic contrast but does not prove that the config atom caused it.
Appendix~\ref{app:trace-xarray} shows the decisive fallback difference and its
official test outcome.

\paragraph{LCB closure and its counterexample.}
On \code{v6-025-abc391-f}, the representative Pi trace writes once and ends
after a failing execution. The recomposed trace writes, encounters two bash
failures, rewrites, executes the test, and receives the task-specific passing
result. This is consistent with a repair-closure mechanism. Across all 1,000
slots, the config-changing recomposition records 20 more trace-strict successes
and 34 fewer schema-shaped errors than Pi, but also 17 more bash-error attempts,
0.171 more tool calls per slot, and about 7.8 seconds more duration per slot.
The Pi-config recomposition records 17 more trace-strict successes, 36 fewer
schema errors, 22 fewer tool-error attempts, 0.042 more tool calls, and about
5.4 seconds more duration per slot. The mixed signs rule out a simple ``more
tools'' mechanism.
On \code{v6-088-arc191-d}, both recompositions regress despite representative
successful traces sharing the same read--read--write--bash shape as Pi; the
available trace categories do not explain the stochastic solution-quality
difference. These trace-based cases support closure and fragility claims, not
the primary clean-positive data claim.

Pi prompt, turn loop, and acceptance are held fixed in the two LCB
recompositions, so those results provide no evidence for replacing those
dimensions. Even among changed dimensions, cases do not isolate a single atom:
session/hooks, provider plugins, tools, optional config, and compatibility
behavior move together. The case set is best understood as mechanistic evidence
for harness-conditioned trajectories and as a source of graded sibling labels.

These paired cases show which supervision can be recovered from sibling
trajectories: clean positives, regressions, semantic near misses, and
preferences between alternative patches. The next analysis checks whether the
export actually contains training records with these distinctions.

\paragraph{Materializing training records.}

From the selected-member SWE validation, a 200-slot slice spanning 20 instances
and ten sibling slots per instance materializes \MetricTrainingRecords{} rows
across several training purposes:

\begin{table}[H]
\centering
\small
\begin{tabularx}{0.82\linewidth}{Y r r}
\toprule
\tablehead Output dataset & Records & Share of 438 \\
\midrule
Clean SFT & 63 & 14.38\% \\
Cleanup-needed SFT & 56 & 12.79\% \\
Critic negatives & 52 & 11.87\% \\
Filter samples & 113 & 25.80\% \\
Preference pairs & 154 & 35.16\% \\
\midrule
\textbf{Total multi-purpose rows} & \textbf{438} & \textbf{100.00\%} \\
\bottomrule
\end{tabularx}
\caption{\textbf{Verified training-data accounting from selected-member SWE
validation.} The export contains 438 derived rows, not 438 independent
rollouts; one verified sibling can contribute to more than one objective.}
\label{tab:training-records}
\end{table}

The 154 preference pairs comprise 46 minimal-cleanup versus larger-cleanup
pairs, 44 clean versus noisy resolved pairs, 32 minimal-clean versus larger
clean pairs, 14 resolved versus regression pairs, 14 resolved versus target
misses, and four resolved versus no-action pairs. The split contains 16 train
instances and four dev instances with no instance crossing the boundary.
The materialized full-mix split contains 339 train and 99 dev rows.

\begin{figure}[H]
  \centering
  \framedgraphic[0.90\linewidth]{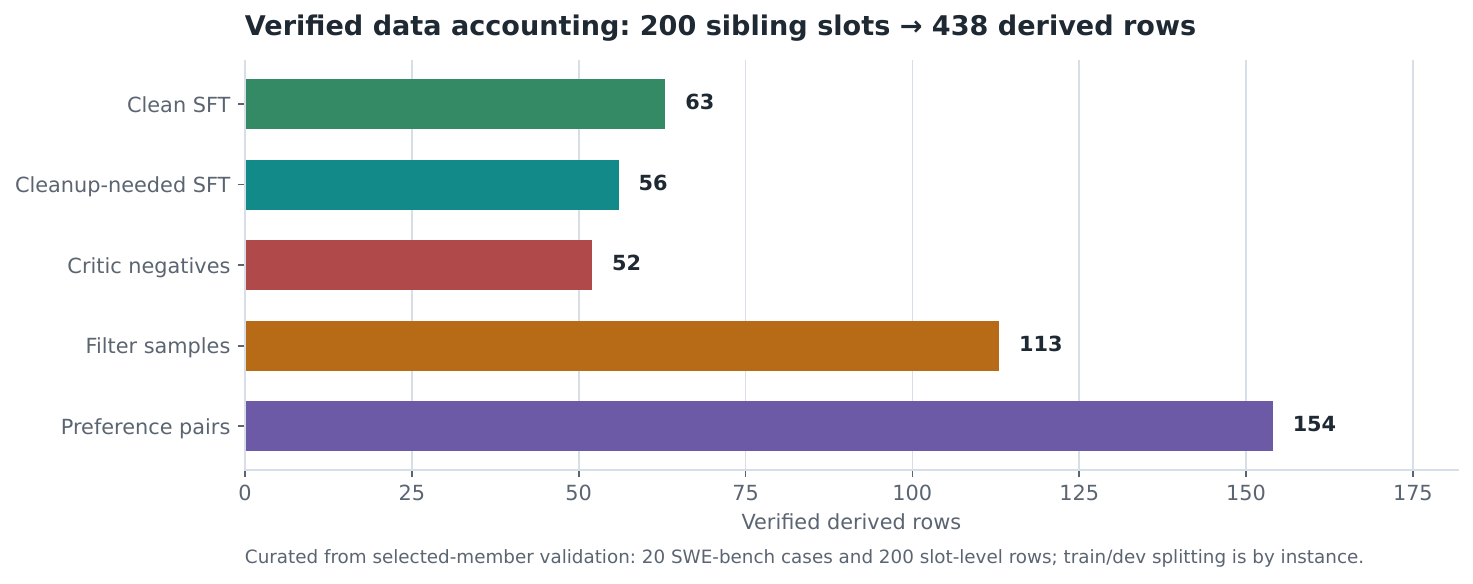}
  \caption{\textbf{Verified multi-purpose rows from the selected-member sibling
  slice.} The audit maps verifier and patch-quality labels to more than clean
  SFT: failed, noisy, and alternative-success siblings supply critic, filter,
  and preference categories.}
  \label{fig:training-data}
\end{figure}

Figure~\ref{fig:training-data} visualizes the same reported row counts by
training purpose.

\begin{findingbox}{Finding 3: sibling contrasts become model-update records}
The distinctions exposed by the paired cases are not only post-hoc
interpretations. The selected-member export maps 200 curated sibling slots into 438 structured
records: successful, failed, noisy, and alternative trajectories become SFT,
critic, filter, and preference data. Harness evolution therefore provides a
concrete data handoff for model updating, rather than only identifying a better
runtime.
\end{findingbox}

Together, Findings 1--3 establish the two outputs of a harness-evolution round:
expanded execution for the current model and verified sibling data for the
update. The discussion reconnects these outputs through build--update--rebuild.

\section[Discussion]{Discussion: From One Evolution Round to Recursive Self-Improvement}
\label{sec:discussion}

\subsection{The return from update to rebuild makes improvement recursive}

The recursion in build--update--rebuild is not recursion over model calls or
subagents within one execution. It is the return of an updated model--harness
state to the same improvement process. Three directed dependencies make that
return explicit: the harness shapes the trajectory
($\phi_t\rightarrow\tau_t$); verified trajectories shape the next model
($\tau_t\rightarrow\theta_{t+1}$); and the updated capability profile changes
the conditions for rebuilding the harness
($\theta_{t+1}\rightarrow\phi_{t+1}$).

The evidence from a round therefore has two destinations. Verified sibling
contrasts provide supervision for the model, while component-level successes,
failures, and incompatibilities inform retention, pruning, and proposal on the
harness side. After the model update, repeating harness evolution under the new
capability profile tests whether the preferred context, tools, recovery
behavior, or stopping policy has moved. A harness that was weak for
$M_{\theta_t}$ may become useful for $M_{\theta_{t+1}}$, while a previously
strong harness may no longer be the best interface.

This view also makes recursive progress measurable. Across rounds, evaluation
should track the model change under a held harness, shifts in harness ranking
under the updated model, changes in the task and failure distribution, and the
amount and quality of verified sibling experience produced by the rebuilt
portfolio. These comparisons can guide the next portfolio, update objective,
and routing policy while keeping model-side and harness-side changes
separately attributable.

\subsection{The model--harness pair has two improvement frontiers}

The two outputs are optimized differently within one round but coupled across
rounds: execution selects a useful runtime for the current model, while the
portfolio retains the contrasts that drive the next update.

The main evolution matrix and its selected-member validations support a
conditional view of observed capability:
\begin{equation}
  \text{observed agent capability}
  =
  f(\text{model},\text{harness},\text{task},\text{budget},\text{verifier}).
\end{equation}
This does not diminish the model; it locates the model inside the system that
turns tokens into environment effects. It also means that harness evolution has
two empirical frontiers rather than one universal winner. The full LCB matrix
contains a 52/100 fixed candidate and a 79/100 portfolio union, while deeper
validation changes which member is preferred. The larger and more consistent
effect is therefore a change in which tasks and trajectories close. That
change has informational value even when the candidate is not selected for
deployment.

The performance frontier rewards a fixed harness that is accurate, reliable,
safe, and affordable. The data frontier rewards a portfolio that produces
novel verified positives, informative failures, clean alternatives, and useful
contrasts. These objectives can disagree: the best deployed harness need not
be the best exploration policy, and a failed candidate can still supply a
well-labeled negative.

\subsection{Portfolio coverage exposes execution headroom and data yield}

Portfolio coverage is not the score of one fixed harness, but it captures two
consequences of rapid evolution.

First, it measures \emph{data yield}. If a candidate portfolio produces a verified
success for a task missed by the baseline, it has created a positive
candidate trajectory even when no online router exists. Outcome, regression,
and hygiene gates must still determine whether it is suitable for
distillation. In this use, compute is spent during data construction rather
than every deployment request.

Second, it bounds the opportunity for routing. The gap between the best fixed
candidate and an oracle union is the maximum improvement available to a
selector over the observed portfolio and attempt budget. Realizing that gap
requires a pre-outcome signal: task features, early trace features, a cheap
critic, or a staged policy that can abandon an unpromising harness. The
current experiments evaluate no such selector, so router performance remains an
open question.

\subsection{Verified failures enrich model-update data}

Under trace-strict status, the 65-candidate matrix contains 23 zero-score
candidates. It would be easy either to discard them as failed engineering or to
retain them indiscriminately as negative training data. Both choices lose
information. In a co-evolution framework, they serve two possible purposes:

\begin{itemize}
  \item They identify invalid or semantically incompatible regions that
  stronger conformance gates should prune before live evaluation.
  \item When the candidate is valid but behaves poorly, its trajectories become
  negatives for compatibility critics, early-failure predictors, or recipe
  proposal policies.
\end{itemize}

However, low-quality runs must not overwhelm training. A harness failure can
teach artifacts of a broken interface rather than robust problem solving; an
apparently successful run can encode a regression or verifier exploit. Data
builders should therefore retain harness identity, verifier rationale, outcome
vector, and failure class so examples can be routed to the right objective,
weighted, or excluded.

\section{Related Work}
\label{sec:related}

The proposed cycle connects four lines of work: recursive self-improvement,
runtime interfaces that shape agent behavior, automated evolution of agent
systems, and learning from environment interaction.

\subsection{Recursive self-improvement and model--harness co-evolution}

Recursive self-improvement spans different objects and degrees of loop closure,
from bounded behavior refinement to changes in model policies, evaluators, and
the research process itself \cite{rsi-survey}. G\"odel Machines provide a strong
self-referential formulation in which a system rewrites its own code after
proving that the change improves utility \cite{goedel-machine}. Darwin G\"odel
Machine replaces proof-based rewriting with empirical variation, benchmark
selection, and an archive of agent-code variants \cite{dgm}. These works treat
the improvement target broadly and motivate asking which state returns to the
next round.

Recent work applies this question directly to harnesses. Recursive Harness
Self-Improvement iteratively revises a prompt-level specification of the agent
loop using pairwise feedback over its revision history \cite{rhi}. HarnessX v3
combines typed, trace-driven harness adaptation with cross-harness GRPO and a
mixed-policy replay buffer in a joint harness--model loop \cite{harnessx}.
Against this established background, \system{} focuses on a complementary
systems layer: recomposing behavior derived from
heterogeneous open-source harnesses while preserving source identity from the
declared recipe through runtime evidence and verifier-labeled sibling records.
This makes the bounded model--harness state transition inspectable rather than
relying on arbitrary self-rewriting.

\subsection{Harnesses as agent--computer interfaces}

ReAct interleaves language reasoning with environment actions
\cite{react}, while Toolformer demonstrates self-supervised learning of when
and how to call external APIs \cite{toolformer}. SWE-agent makes the interface
itself an experimental object: its agent--computer interface changes how a
model navigates repositories, edits files, and runs tests
\cite{sweagent}. These works establish that agent behavior depends on more than
static next-token prediction. \system{} extends the intervention surface from
prompt/tool use to the complete runtime in
Equation~\ref{eq:harness-tuple}, including session semantics, configuration,
turn control, acceptance, and policy. We build on this premise rather than
claiming priority for treating runtime design as part of the agent.

\subsection{Automated harness and workflow evolution}

AutoFlow generates natural-language agent workflows and iteratively optimizes
them \cite{autoflow}. Automated Design of Agentic Systems (ADAS) proposes
automatically evolving agent designs and demonstrates Meta Agent Search, where a meta
agent writes new code-defined agents \cite{adas}. AFlow casts code-represented
workflow optimization as Monte Carlo tree search \cite{aflow}. These systems
report the quality of discovered workflows or agent programs. Our evaluation
additionally treats trajectories from non-selected candidates as an explicit
output linked to matched siblings and graded execution evidence. A non-winning
candidate can therefore contribute a regression example, semantic near miss,
artifact-quality negative, or preference contrast.

\subsection{Interaction data and executable software benchmarks}

LiveCodeBench provides a continuously updated evaluation for code capabilities
\cite{livecodebench}; SWE-bench turns real GitHub issues into executable
repository-level tasks \cite{swebench}. SWE-Gym supplies software-engineering
training environments and trajectories for agents and verifiers
\cite{swegym}, while SWE-smith scales synthetic software-engineering task
construction across repositories \cite{swesmith}. Learn-by-Interact constructs
agent training data from environment interaction trajectories and applies it
to training and in-context adaptation \cite{learnbyinteract}.

Taken together, prior work improves runtime interfaces, evolves workflows, or
learns from interaction data. \system{} connects these strands through a
source-traceable build-to-update handoff. The same declared interventions yield
evidence about current execution and matched sibling records for subsequent
model updating, making the connection explicit and auditable at the level of
source-derived harness components and matched sibling records.

\section{Limitations and Threats to Validity}
\label{sec:limitations}

\subsection{Scope of the evaluation}

The main evolution matrix uses 100 AtCoder tasks and checks only their public tests.
It therefore does not measure hidden-test or official LiveCodeBench performance.
The tasks are the first 100 eligible examples rather than a random sample. The
selected-member SWE follow-up covers 55 Verified tasks. All evaluations concern
coding agents, so the results may not apply to other types of agents.

\subsection{What causes an improvement}

Each evolved harness changes several components at once. We can measure the
effect of the whole harness, but we cannot tell which individual component
caused the change. Future experiments should change one component at a time
while keeping the others fixed.

\subsection{Effects of attempts and selection}

Coverage usually increases when more attempts or harnesses are tried. Oracle
coverage is calculated after seeing all outcomes, so it does not represent a
harness choice made before deployment. The 65-candidate matrix uses one attempt
per candidate--task pair, while the selected-member LCB and SWE validations use
ten and two attempts, respectively. Their rankings are therefore not directly
comparable, and selecting members for deeper evaluation can introduce
selection effects. Future work should use separate tasks for evolution and
final evaluation, repeat every candidate under a common budget, and compare
methods under the same total compute.

\subsection{Limits of automated evaluation}

Tests do not capture every requirement. A patch may pass while changing tests,
making unnecessary edits, or exploiting the evaluator. A correct patch may
also fail because of a timeout or a broken environment. We keep several labels
to make these cases visible, but some still require human review.

\subsection{Data scale and model updating}

This paper materializes model-update data but does not train an updated model.
First, the verified sibling dataset must be scaled beyond the current
evaluation before model updating can be evaluated convincingly. \system{} can
automate candidate execution, verification, and data export at larger scale,
but generating broad portfolios and repeated rollouts still requires substantial
token budgets. Second, closing the loop requires more than additional data: it
also requires training methods suitable for RSI that can learn from
heterogeneous sibling signals while controlling regressions and preserving
general capability across rounds. Both cost-efficient data scaling and
reliable recursive model updating remain important directions for completing
multi-round build--update--rebuild experiments.

\subsection{Safety and maintenance}

Agents can run unsafe commands, change files, and leak secrets through traces.
Combining components from different harnesses can also create unsafe
permissions. Experiments should therefore use isolated workspaces, limited
permissions and network access, and trace redaction. Because the upstream
harnesses change over time, their integrations and safety checks must be tested
again after an update.

\section{Conclusion}
\label{sec:conclusion}

Recursive self-improvement for an interactive agent should improve the
executed system, not the model in isolation. At round $t$, the harness
portfolio $\Phi_t$ changes which behaviors $M_t$ can reach and which
experiences it produces. The resulting verified experience can update the
model to $M_{t+1}$; because that update changes the capability profile, it can
also change which harness is preferred. The recursively improving state is therefore
$Z_t=(M_t,\Phi_t)$, and the return from $Z_t$ to $Z_{t+1}$ is the defining
structure of model--harness co-evolution.

Model, harness, and data therefore form a coupled improvement system. The
model's capability profile determines which harnesses are effective; each
harness changes how the model acts and which experiences it produces; and
those experiences become the data that shapes the next model. \system{} makes
this three-way influence usable for RSI by evolving harnesses around the
current model and preserving verified differences among their trajectories as
model-update data. Harness evolution then serves not only current execution but
also the next round of system improvement.

Build--update--rebuild turns this three-way influence into an RSI process.
Harness evolution produces verified experience; model training absorbs that
experience; and the updated model returns to harness evolution rather than
inheriting a permanently fixed runtime. Scaling experience, applying reliable
model updates, and comparing held with rebuilt harnesses across rounds are the
next steps toward repeated $Z_t\rightarrow Z_{t+1}$ improvement. \system{}
makes this recursive transition auditable.

\clearpage
\appendix

\section{Selected-Member Follow-up Results}
\label{app:selected-results}

Every candidate below is already a member of the 65-candidate evolution
portfolio. These tables report their deeper evaluation.

\subsection{Repeated LCB evaluation}

\begin{table}[H]
\centering
\small
\begin{tabularx}{\linewidth}{Y r r}
\toprule
\tablehead Selected member & Tasks $\geq$1/10 & Successful slots \\
\midrule
Pi Mono & 72/100 & 529/1,000 \\
OpenCode session/hooks/config/tools; other modules Pi &
\textbf{75/100} & \textbf{549}/1,000 \\
OpenCode session/hooks/tools; other modules Pi &
\textbf{75/100} & 546/1,000 \\
\bottomrule
\end{tabularx}
\caption{\textbf{Repeated-run validation of selected portfolio members.}
All LCB outcomes use trace-strict status. Tasks count at least one successful
attempt out of ten; slots count successful attempts out of 1,000. ``Pi core''
denotes Pi prompt, turn loop, and acceptance; the third row also retains Pi
config.}
\label{tab:lcb-fixed}
\end{table}

\subsection{Official SWE-bench evaluation}

\begin{table}[H]
\centering
\footnotesize
\begin{tabularx}{\linewidth}{Y r r r r r}
\toprule
\tablehead Selected member & Nonempty & Resolved slots & Gaps & Empty &
Tasks $\geq$1/2 \\
\midrule
Pi Mono & 108/110 & 81 & 27 & 2 & 44/55 \\
Hermes session/hooks + Pi remainder & 108/110 & 81 & 27 & 2 & 44/55 \\
Hermes session/hooks/config + Nanobot tools + Pi core &
109/110 & 80 & 29 & 1 & 44/55 \\
OpenCode session/hooks/tools + Pi config/core &
109/110 & 83 & 26 & 1 & \textbf{46/55} \\
OpenCode session/hooks/config/tools + Pi core &
109/110 & \textbf{86} & 23 & 1 & \textbf{46/55} \\
\midrule
\multicolumn{5}{l}{\textit{Selected-member post-hoc union}} &
\textit{49/55} \\
\bottomrule
\end{tabularx}
\caption{\textbf{Official-evaluator validation of selected portfolio
members.} Each member has two slots per task. ``Resolved slots'' counts
successful attempts; the final column counts tasks with at least one resolved
attempt. The union is portfolio coverage, not fixed-harness performance.}
\label{tab:swe-results}
\end{table}

\clearpage
\section{Verifier-Grounded Case Traces}
\label{app:case-traces}

The following tables reconstruct the two paired cases from the persisted
rollout transcript, extracted patch, and official SWE-bench evaluator output.
They retain the chronological, decision-relevant steps---inspection, edit,
in-rollout checks, and official validation---while omitting repeated file reads
and repeated model narration.

\subsection{Pytest: target success with regressions}
\label{app:trace-pytest}

Both runs identify the missing serialization of chained exceptions and pass
the two target tests. Their patches differ in backward compatibility: the Pi
patch replaces the existing \code{longrepr} schema for chained exceptions,
whereas the evolved sibling extends it. Table~\ref{tab:trace-pytest} exposes
where that difference enters the trace and how the official evaluator changes
the label.

\begin{table}[H]
\centering
\scriptsize
\begin{tabularx}{\linewidth}{
  >{\RaggedRight\arraybackslash}p{0.15\linewidth}
  >{\RaggedRight\arraybackslash}p{0.39\linewidth}
  Y}
\toprule
\tablehead Trace stage & Pi baseline, attempt 1 & Evolved sibling, attempt 1 \\
\midrule
Run identity &
\code{pi-mono}; 43 observed tool calls in 355 seconds. &
OpenCode session/hooks/config/tools with Pi prompt/turn; 54 observed tool calls
in 377 seconds. \\
Inspection &
Searches for report serialization, then repeatedly reads
\path{src/_pytest/reports.py} and \path{src/_pytest/_code/code.py}. It correctly
identifies \code{ExceptionChainRepr.chain} as the missing information. &
Reads the same implementation files and also inspects
\path{testing/test_reports.py}, including the existing serialization contract
that accesses top-level \code{reprtraceback}. \\
Patch &
Changes only \path{src/_pytest/reports.py} (6,400 bytes). For a chained
exception, it emits a new \code{longrepr} object containing only
\code{chain} and \code{sections}. &
Changes \path{src/_pytest/_code/__init__.py} and
\path{src/_pytest/reports.py} (6,854 bytes). It exports
\code{ExceptionChainRepr}, preserves \code{reprcrash},
\code{reprtraceback}, and \code{sections}, and adds the full \code{chain}. \\
In-rollout checks &
Writes a custom chained-exception script and observes successful round trips.
Attempts to run repository tests stop at the local environment guard
\code{requires pytest-2.0}; the compatibility tests are therefore not closed
inside the rollout. &
Attempts \path{testing/test_reports.py} and encounters the same environment
guard, then performs custom import and chain round-trip checks. These checks
exercise the new path but do not substitute for the official regression set. \\
Official targets &
Both FAIL\_TO\_PASS tests pass (2/2). &
Both FAIL\_TO\_PASS tests pass (2/2). \\
Official regressions and label &
121/123 PASS\_TO\_PASS tests pass. The two failures,
\path{test_xdist_longrepr_to_str_issue_241} and
\path{test_deserialization_failure}, raise \code{KeyError: reprtraceback}.
Label: \code{gaps-found}; data role: regression-aware critic negative. &
123/123 PASS\_TO\_PASS tests pass. Label: \code{resolved}; data role: clean
positive and preference winner. \\
\bottomrule
\end{tabularx}
\caption{\textbf{Condensed Pytest paired trace.} Both patches solve the target,
but only the evolved sibling preserves the pre-existing serialization schema.}
\label{tab:trace-pytest}
\end{table}

\begin{codeblock}[Decisive Pytest schema difference]
Pi:      longrepr = {"chain": chain, "sections": sections}
Sibling: longrepr = {"reprcrash": reprcrash,
                     "reprtraceback": reprtraceback,
                     "sections": sections,
                     "chain": chain}
\end{codeblock}

The official regression failures are therefore not incidental test noise. They
directly query a field removed by the Pi patch, even though that patch already
passes both newly introduced target tests. This is why the paired trajectories
form a useful resolved-versus-regression supervision pair.

\subsection{Xarray: a clean semantic near miss}
\label{app:trace-xarray}

The Xarray pair holds the OpenCode session/hooks/tools, Pi prompt/turn, and
attempt index fixed while changing the config atom. Both patches are clean,
single-file edits to the same expression, and both pass their ad hoc checks.
Table~\ref{tab:trace-xarray} shows why their official labels nevertheless
diverge.

\begin{table}[H]
\centering
\scriptsize
\begin{tabularx}{\linewidth}{
  >{\RaggedRight\arraybackslash}p{0.15\linewidth}
  >{\RaggedRight\arraybackslash}p{0.39\linewidth}
  Y}
\toprule
\tablehead Trace stage & Pi-config sibling, attempt 2 & OpenCode-config sibling, attempt 2 \\
\midrule
Run identity &
24 observed tool calls in 130 seconds; 837-byte patch. &
27 observed tool calls in 122 seconds; 778-byte patch. \\
Diagnosis &
Locates \code{where()} in \path{xarray/core/computation.py}, traces attribute
collection through \code{apply\_ufunc}, and reproduces the
\code{IndexError} for scalar \code{x} with \code{keep\_attrs=True}. &
Locates the same expression and reproduces the same \code{IndexError}. \\
Patch &
If \code{attrs[1]} is unavailable, falls back to \code{attrs[0]} when present,
which can be the condition's attributes. &
If \code{attrs[1]} is unavailable, returns an empty dictionary because scalar
\code{x} has no attributes to preserve. \\
In-rollout checks &
Runs several direct Python examples and reports them as passing; its scalar
example uses an attribute-empty condition. The repository test command cannot
run because that interpreter has no \code{pytest} module. &
Runs direct Python examples covering scalar and DataArray inputs and reports
them as passing. The repository test command encounters the same missing
\code{pytest} module. \\
Official target &
The added \path{test_where_attrs} case uses condition attributes
\code{\{'attr': 'cond'\}} and scalar \code{x}. The patch returns those condition
attributes instead of \code{\{\}}, so FAIL\_TO\_PASS is 0/1. &
The same test receives \code{\{\}} as required, so FAIL\_TO\_PASS is 1/1. \\
Official regressions and label &
247/247 PASS\_TO\_PASS tests pass. Label: \code{gaps-found}; data role: clean
semantic negative. &
247/247 PASS\_TO\_PASS tests pass. Label: \code{resolved}; data role: clean
positive and preference winner. \\
\bottomrule
\end{tabularx}
\caption{\textbf{Condensed Xarray paired trace.} Patch hygiene and local
examples are matched; the official target isolates the fallback semantic that
separates the near miss from the resolved patch.}
\label{tab:trace-xarray}
\end{table}

\begin{codeblock}[Decisive Xarray fallback difference]
Gap:      attrs[1] if len(attrs) > 1 else
          (attrs[0] if len(attrs) == 1 else {})
Resolved: attrs[1] if len(attrs) > 1 else {}
\end{codeblock}

This pair explains why patch cleanliness alone is insufficient as a training
label. The negative is structurally clean and preserves every regression test,
yet encodes the wrong fallback semantics for the target behavior.

\end{document}